\documentclass[%
superscriptaddress,
preprint,
 amsmath,amssymb,
 aps,
]{revtex4-1}

\usepackage{graphicx}
\usepackage{dcolumn}
\usepackage{bm}
\usepackage{xcolor}
\usepackage{qcircuit}
\usepackage{braket}
\usepackage{hyperref}

\renewcommand{\figureautorefname}{Figure~\negthinspace}

\begin{document}

\preprint{BNL-SBU}

\title{Quantum Convolutional Neural Networks for High Energy Physics Data Analysis}

\author{Samuel Yen-Chi Chen}
\email{ychen@bnl.gov}
\affiliation{Computational Science Initiative, Brookhaven National Laboratory, Upton, NY 11973, USA}%

\author{Tzu-Chieh Wei}
\email{tzu-chieh.wei@stonybrook.edu}
\affiliation{C. N. Yang Institute for Theoretical Physics and Department of Physics and Astronomy, State University of New York at Stony Brook, Stony Brook, NY 11794-3840, USA}

\author{Chao Zhang}
\email{czhang@bnl.gov}
\affiliation{Physics Department, Brookhaven National Laboratory, Upton, NY 11973, USA}

\author{Haiwang Yu}
\email{hyu@bnl.gov}
\affiliation{Physics Department, Brookhaven National Laboratory, Upton, NY 11973, USA}

\author{Shinjae Yoo}%
 \email{sjyoo@bnl.gov}
\affiliation{Computational Science Initiative, Brookhaven National Laboratory, Upton, NY 11973, USA}




\date{\today}

\begin{abstract}
This work presents a quantum convolutional neural network (QCNN) for the classification of high energy physics events. The proposed model is tested using a simulated dataset from the Deep Underground Neutrino Experiment. The proposed architecture demonstrates the quantum advantage of learning faster than the classical convolutional neural networks (CNNs) under a similar number of parameters. In addition to faster convergence, the QCNN achieves greater test accuracy compared to CNNs. Based on experimental results, it is a promising direction to study the application of QCNN and other quantum machine learning models in high energy physics and additional scientific fields.

\end{abstract}

\maketitle


\section{\label{sec:Indroduction}Introduction}

High Energy Physics (HEP) communities have a long history of working with large data and applying advanced statistics techniques to analyze experimental data in the energy, intensity, and cosmic frontiers. With ever-increasing data volumes, the HEP community needs a significant computational breakthrough to continue this trajectory, and tools developed in Quantum Information Science (QIS) could provide a viable solution. Quantum advantage is the potential to solve problems faster than any classical methods \cite{arute2019quantum, harrow2017quantum}. In computational-complexity-theoretic terms, this generally means providing a superpolynomial speedup over the best known or possible classical algorithm \cite{nielsen2002quantum}.

Machine learning methods promise great benefits for scalable data analytics. The big wave of deep learning algorithm development stems
from recent advances in convolutional neural networks (CNNs) \cite{lecun1998gradient}, which can effectively capture spatial dependencies within an image, as well as automatically learn important features from them \cite{krizhevsky2012imagenet, szegedy2015going, simonyan2014very, lecun2015deep, goodfellow2016deep}.
Along with big data and graphics processing unit (GPU) processing capabilities, deep learning has significantly improved the ability to analyze large volumes of images. There are several examples 
where CNNs have been successfully applied to HEP challenges using classical computers~\cite{hep-ml}. 
However, in quantum computing, no significant progress has been
made toward implementing such robust representation learning methods to date.

In this work, we present a new hybrid Quantum Convolutional Neural Network (QCNN) framework to demonstrate the quantum advantage versus corresponding classical algorithms. We simulate its performance for classification of HEP events from the simulated data in neutrino experiments. We show that with a similar number of parameters in QCNN and classical CNNs, the QCNN can learn faster or reach better testing accuracy with fewer training epochs. Thus, our simulations empirically demonstrate the quantum advantage of QCNN over CNN in terms of testing accuracy.
This paper is organized as follows: Section~\ref{sec:hep-data} introduces the HEP experimental data used in this work. In Section~\ref{sec:vqc} and \ref{sec:qcnn}, we describe the new QCNN architecture in detail. Section~\ref{sec:results} shows the performance of the QCNN on the experimental data. Section~\ref{sec:conclusion} concludes the overall discussion.
%
\section{High Energy Physics Data} \label{sec:hep-data}
\begin{figure}[htbp]
\centering
\includegraphics[width=0.85\linewidth]{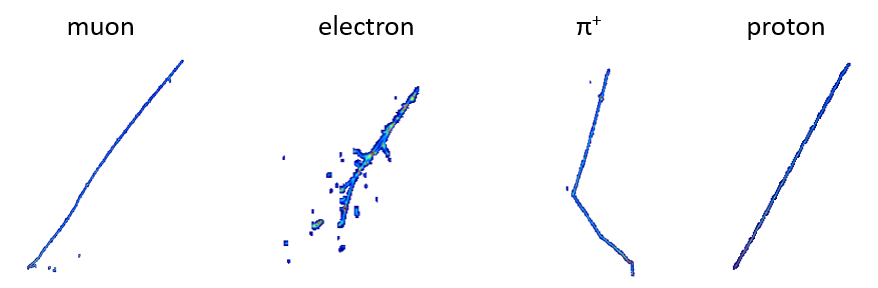}
\caption{Example images of simulated particle activities ($\mu^+$, $e^-$, $\pi^+$, $p$) in a LArTPC detector. Colors in the images represent the sizes of the ionization energy loss along the particle trajectories when measured by LArTPC's wire planes.}
\label{simu_particle}
\end{figure}

In this work, we use simulated data from the Deep Underground Neutrino Experiment (DUNE)~\cite{dune} to develop and test our QCNN algorithms on high-energy experiments. Hosted in the United States, DUNE is the next-generation, international, world-class experiment~to reveal new symmetries of nature. 
DUNE's primary goals include searching for CP violation in the lepton sector, determining neutrino mass ordering, performing precision tests of the three-neutrino paradigm, detecting supernova neutrino bursts, and searching for nucleon decays beyond the Standard Model.  
DUNE is an excellent test case for the QCNN because its main detector technology, the Liquid Argon Time Projection Chamber (LArTPC), effectively provides high-resolution images of particle activities as the ionized electrons drift toward the multiple sensing wire planes~\cite{rubbia77,Chen:1976pp,willis74,Nygren:1976fe}. 
An advanced LArTPC simulation package, the Wire-Cell Toolkit~\cite{wct} and LArSoft software~\cite{larsoft}, is used to generate realistic single-particle images in a LArTPC detector. The simulation implements a chain of algorithms, including: 1) generating single-particle kinematics, 2) applying LArTPC detector response, 3) adding realistic electronic noise, and 4) performing digital signal processing. Details about the LArTPC simulation can be found in Ref~\cite{lartpc-sp}. 
Four different types of particles ($\mu^+$, $e^-$, $\pi^+$, and $p$) are simulated. \figureautorefname{\ref{simu_particle}} shows sample images of simulated particle activities on the collection wire plane. The images have a resolution of 480 x 600 pixels, where each pixel in the x-axis represents a single wire and each pixel in the y-axis represents a sampling time tick. 
In this work, the goal of the QCNN is to predict the types of different particles by analogy with those performed via the classical CNN~\cite{lartpc-cnn}. Each particle's momentum is set such that the mean range of the particle is about 2 meters, so the classification is not sensitive to the image size.

As visualized in~\figureautorefname{\ref{simu_particle}}, the classification of the four different particles primarily is a pattern recognition problem. A positively charged muon ($\mu^+$) is a track-like particle, while an electron ($e^-$) produces electromagnetic showers that are spatially extended. A muon is a minimum ionizing particle in terms of energy loss along its trajectory, which translates into the intensity of the pixels. It experiences Multiple Coulomb Scattering (MCS) when passing the detector, causing its trajectory to deviate from a straight line. It also decays into a low-energy positron after it loses most of the kinetic energy and stops in the detector, leading to another short track segment near the end of the main track. A positively charged pion ($\pi^+$) looks similar to a muon in terms of energy loss, MCS, and decay, but it experiences additional nuclear interactions during its passage in the detector, often leading to a hard scattering (represented as a ``kink'') along its main trajectory. Finally, a proton ($p$) also is a track-like particle. However, because a proton's mass is much heavier than a muon or pion, it has higher energy loss and encounters less MCS during travel. Consequently, a proton's track has higher intensity and is straighter than those from muons and pions. 

These diverse features in detector images make the LArTPC data analysis well suited for CNN-type machine learning algorithms rather than hand-crafted feature extraction methods. Previous work with LArTPC has shown excellent performance from single-particle classification~\cite{lartpc-cnn} to the more complicated neutrino interaction classifications~\cite{dune-cnn}. In this work, we perform a quantum implementation of the classical CNN through the variational quantum circuits for the first time in LArTPC data analysis. By comparing the performance to the classical CNN, we explore possible quantum acceleration and advantage in machine learning for HEP data analysis. 

\section{Variational Quantum Circuits}\label{sec:vqc}
\begin{figure}[htbp]
\begin{center}
\begin{minipage}{10cm}
\Qcircuit @C=1em @R=1em {
\lstick{\ket{0}} & \multigate{3}{F(\mathbf{x})}  & \qw        & \multigate{3}{V(\boldsymbol{\theta})}       & \qw      & \meter \qw \\
\lstick{\ket{0}} & \ghost{F(\mathbf{x})}         & \qw        & \ghost{V(\boldsymbol{\theta})}              & \qw      & \meter \qw \\
\lstick{\ket{0}} & \ghost{F(\mathbf{x})}         & \qw        & \ghost{V(\boldsymbol{\theta})}              & \qw      & \meter \qw \\
\lstick{\ket{0}} & \ghost{F(\mathbf{x})}         & \qw        & \ghost{V(\boldsymbol{\theta})}              & \qw      & \meter \qw \\
}
\end{minipage}
\end{center}
\caption[General structure for the variational quantum circuit.]{{\bfseries General structure for the variational quantum circuit (VQC).}
The $F(\mathbf{x})$ is the quantum operation for encoding the classical data into the quantum state and $V(\boldsymbol{\theta})$ is the variational quantum circuit block with the adjustable parameters $\boldsymbol{\theta}$.
}
\label{Fig:GeneralVQC}
\end{figure}
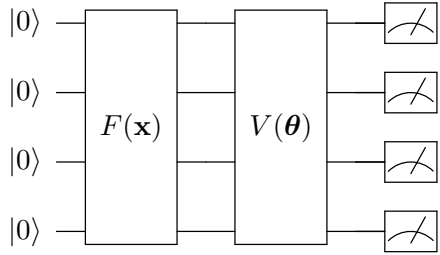

Variational quantum circuits (VQC) are quantum circuits that have \emph{tunable} or \emph{adjustable} parameters subject to classical iterative optimizations, which are commonly based on gradient descent and its variants \cite{schuld2019evaluating, benedetti2019parameterized}. The general structure of VQC is presented in~\figureautorefname{\ref{Fig:GeneralVQC}}. Here, the $F(\mathbf{x})$ block is for the state preparation that encodes the classical data $\mathbf{x}$ into the quantum state for the circuit to operate on and is not subject to optimization. This state preparation part is designed according to the given research problem. 
The $V(\boldsymbol{\theta})$ block represents the variational or \emph{learning} part. The \emph{learnable} parameters labeled with  $\boldsymbol{\theta}$ will be optimized through gradient-based methods. For example, the commonly used gradient-based optimizers are Adam \cite{kingma2014adam} and RMSProp \cite{Tieleman2012}. In concept, these parameters are comparable to the \emph{weights} in classical deep neural networks (DNNs).
In the final part of this VQC block, we perform the quantum measurement on a subset (or all) of the qubits to retrieve the information. If we run the circuit once and perform a single quantum measurement, it will yield a bit string, such as $0010$, and it generally differs from what we will get if we prepare the circuit again and perform another quantum measurement due to the stochastic nature of quantum systems. However, if we prepare the same circuit and perform the quantum measurement several times, e.g., $1000$ times, we will get the expectation values on each qubit, which should be quite close to the results from theoretical calculation. For example, consider a two-qubit system $\ket{\Psi}$, in every single measurement, the result is one of the following: $00$, $01$, $10$, and $11$. If we prepare $\ket{\Psi}$ and measure it $1000$ times, we will get numerous $00, \cdots ,11$. We can count the frequencies of the appearance of $0$ and $1$ in each qubit and use them to estimate the expectation values. For example, after $1000$ times of repeated measurement, we get $600$ times of $0$ and $400$ times of $1$ in the first qubit. Therefore, the expectation value of the first qubit is $0.4$. In an $N$-qubit system, we place the expectation values of all qubits into a $N$-dimensional vector, which can be processed further in classical or quantum neural networks. We may choose different bases for the measurement.
For example, in this work, we exclusively use the Pauli-$Z$ expectation values at the end of VQC. Although VQCs are simple in concept, they are successful in machine learning tasks. Recent studies have reported the application of such variational architectures in the field of classification \cite{mitarai2018quantum, schuld2018circuit, Farhi2018ClassificationProcessors, benedetti2019parameterized, mari2019transfer, abohashima2020classification, easom2020towards, sarma2019quantum, stein2020hybrid,chen2020hybrid}, function approximation \cite{chen2020quantum, mitarai2018quantum,kyriienko2020solving}, generative machine learning \cite{dallaire2018quantum, stein2020qugan, zoufal2019quantum, situ2018quantum,nakaji2020quantum}, metric learning \cite{lloyd2020quantum, nghiem2020unified}, deep reinforcement learning \cite{chen19, lockwood2020reinforcement, jerbi2019quantum, Chih-ChiehCHEN2020}, sequential learning \cite{chen2020quantum, bausch2020recurrent}, and speech recognition \cite{yang2020decentralizing}. 
\section{Quantum CNN}\label{sec:qcnn}
\emph{CNNs} have been tremendously successful in a wide spectrum of modern machine learning tasks, especially in the area of computer vision \cite{krizhevsky2012imagenet, szegedy2015going, simonyan2014very, lecun2015deep, lecun1998gradient, goodfellow2016deep}. Such methods also afford new insights and progress in scientific research, for example, in HEP event classification \cite{lartpc-cnn, aurisano2016convolutional} and phase transition studies \cite{tanaka2017detection}. With recent advances in quantum computing hardware \cite{preskill2018quantum, cross2018ibm, arute2019quantum}, it is interesting to study the potential advantages and application scenarios for CNNs in the quantum regime.
\emph{QCNN} is the framework that uses VQCs to perform the convolutional operations.
In this work, we replace the classical neural-network-based convolutional filters, or \emph{kernels}, with VQCs to harvest the expressive power granted by quantum entanglements.
The quantum convolutional kernels will sweep through the input image pixels and transform them into a \emph{representation vector} of lower dimensions by performing measurements (see~\figureautorefname{\ref{QCNN_Operation}}).

A stack of VQCs will ensure features of varied length scales are captured in different layers.
\subsection{Quantum Convolutional Filters}
This section describes the VQC building blocks for the QCNN architecture. 
\subsubsection{Data Encoding Layer}
In this layer, we first \emph{encode} a classical input vector into a quantum state, which is necessary for additional processing.
A general $N$-qubit quantum state can be represented as:
\begin{equation}
\label{eqn:quantum_state_vec}
    \ket{\psi} = \sum_{(q_1,q_2,...,q_N) \in \{ 0,1\}^N}^{} c_{q_1,...,q_N}\ket{q_1} \otimes \ket{q_2} \otimes \ket{q_3} \otimes ... \otimes \ket{q_N},
\end{equation}
where $ c_{q_1,...,q_N} \in \mathbb{C}$ is the \emph{amplitude} of each quantum state and $q_i \in \{0,1\}$. 
The square of the amplitude $c_{q_1,...,q_N}$ is the \emph{probability} of measurement with the post-measurement state in  $\ket{q_1} \otimes \ket{q_2} \otimes \ket{q_3} \otimes ... \otimes \ket{q_N}$, and the total probability should sum to $1$, i.e.,
\begin{equation} 
\label{eqn:quantum_state_vec_normalization_condition}
\sum_{(q_1,q_2,...,q_N) \in \{ 0,1\}^N}^{} ||c_{q_1,...,q_N}||^2 = 1. 
\end{equation}
In the proposed framework, the input to the VQC is a matrix with a dimension $n \times n$, where $n$ is the filter/kernel size. The input will first be flattened and transformed into \emph{rotation angles} for the quantum gates. 
In general, the input values of pixels are not in the interval of $[-1, 1]$. We use the arc tangent function to transform these input values into rotation angles. 
For each of the $x_i$ in the $n \times n$ input, there will be two rotation angles generated, $\arctan(x_i)$ and $\arctan(x_i^2)$. The ($n \times n$)-dimensional vector will then be transformed into $2\times n \times n$ angles for the single-qubit rotation.
\subsubsection{Variational Layer}
After encoding the classical values into a quantum state, it will be subject to a series of unitary transformations.
The variational layer (grouped in a dashed-line box in~\figureautorefname{\ref{Fig:Basic_VQC}}) consists of two parts. One is the \emph{entanglement part}, which is a group of CNOT gates. The other is the \emph{rotation part} that includes several single-qubit unitary rotations parameterized by $3$ parameters $\alpha_i$, $\beta_i$, and $\gamma_i$, where $i$ represents the index of qubits. The parameters labeled by $\alpha_i$, $\beta_i$, and $\gamma_i$ are the ones that will be updated by the optimization procedure.
\subsubsection{Quantum Measurement Layer}
To obtain the transformed data from VQC blocks, we perform quantum measurements. 
We consider the ensemble samplings (expectation values) of the VQCs. If working with quantum simulation software (for example, PennyLane~\cite{bergholm2018pennylane} or IBM Qiskit), this value can be calculated deterministically. While implementing on a real quantum computer, it is required to prepare the same system and carry out the measurements repeatedly to gain enough statistics. 
In the proposed QCNN architecture, the quantum convolutional filter will output a single value for each sweep step. Here, we perform the quantum measurements on the first qubit to get the expectation value.
\subsection{Quantum Convolutional Operations}
\begin{figure}[htbp]
\centering
\includegraphics[width=1.0\linewidth]{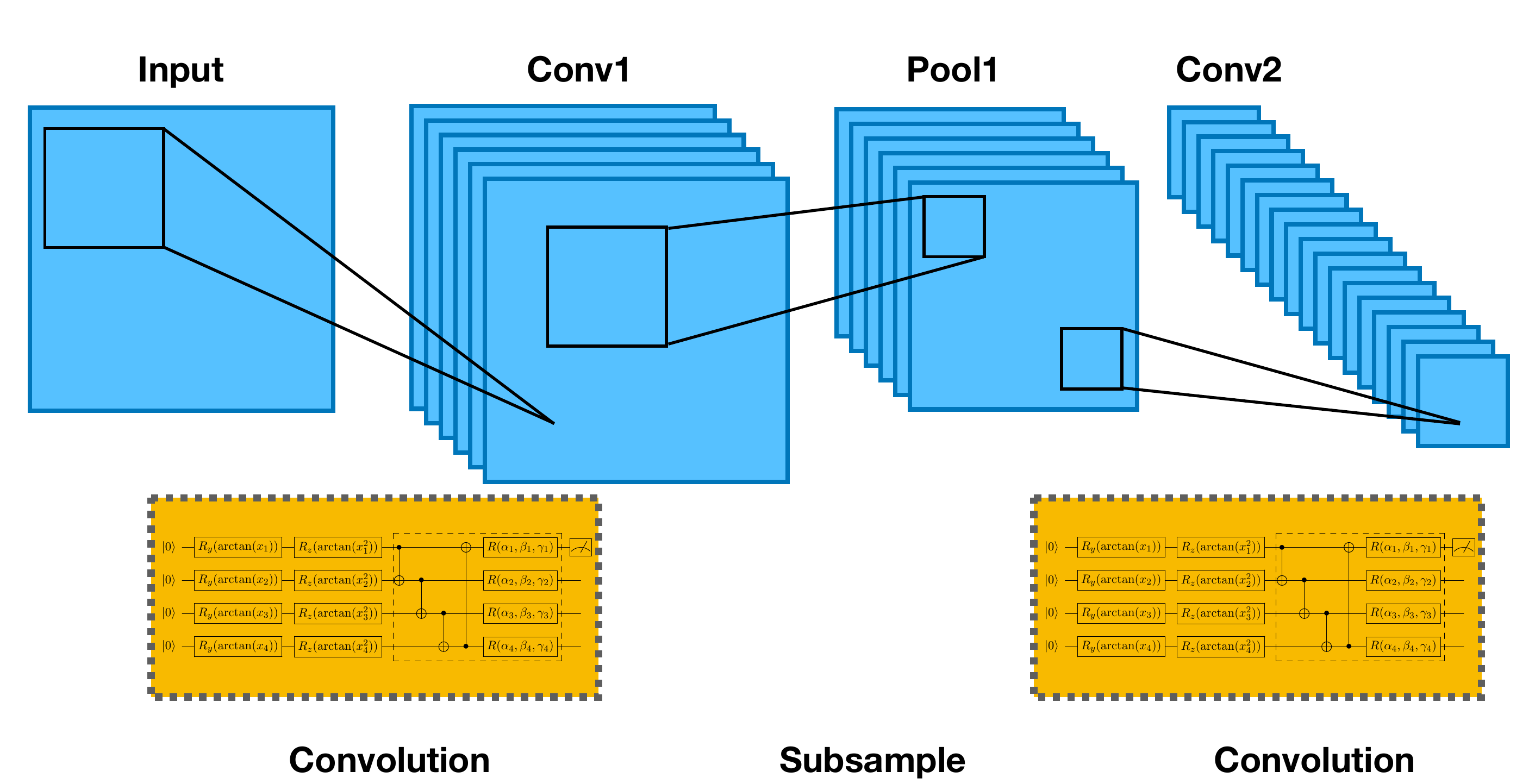}
\caption[Concepts: Quantum CNN.]{{\bfseries Quantum CNN Architecture.} In the proposed hybrid quantum-classical model, the \emph{filter} or \emph{kernel} is a variational quantum circuit as shown in \figureautorefname{\ref{Fig:Basic_VQC}}. Classical pooling and nonlinear activation functions can be optionally added between the convolutional layers analogous to the classical CNN.}
\label{QCNN_Concept}
\end{figure}

\begin{figure}[htbp]
\centering
\includegraphics[width=1.0\linewidth]{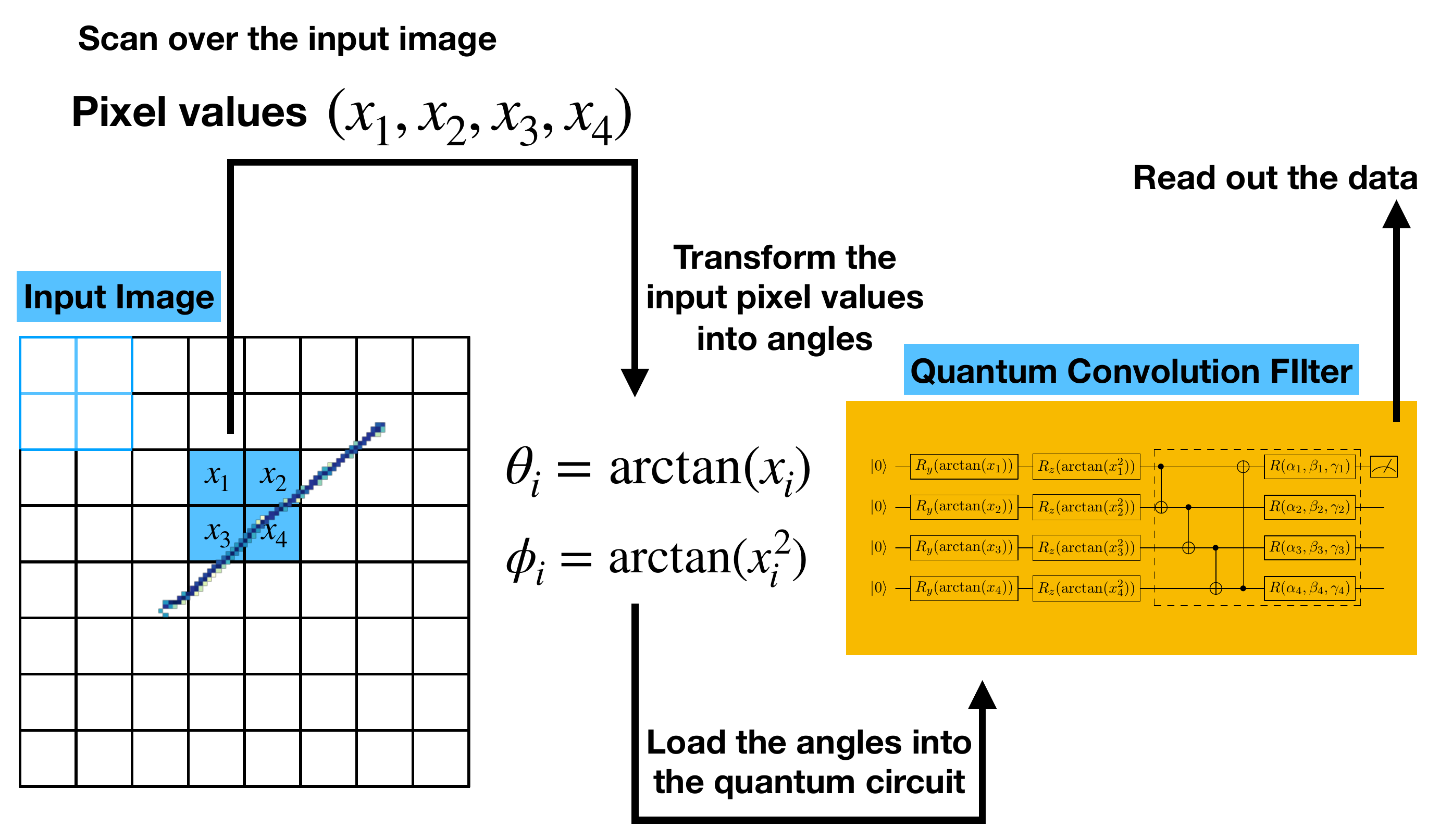}
\caption[QCNN Operation.]{{\bfseries QCNN Operation.} In the QCNN operation, the input  pixel values $(x_1, x_2, x_3, x_4)$ will first be encoded into a quantum state via the variational encoding method. Each value $x_i$ is mapped into two values $\arctan(x_i)$ and $\arctan(x_i^2)$ for the $R_y$ and $R_z$ rotation angles, respectively. The quantum gates parameterized by $\alpha_i, \beta_i, \gamma_i$ then act on this encoded state. At the end of the circuit, Pauli-$Z$ expectation values are retrieved. The retrieved values can then be processed with another layer of quantum convolutional layer or other classical operations (e.g. pooling, nonlinear activation functions or dropout). }
\label{QCNN_Operation}
\end{figure}

\begin{figure}[htbp]
\begin{center}
\begin{minipage}{10cm}
\Qcircuit @C=1em @R=1em {
\lstick{\ket{0}} & \gate{R_y(\arctan(x_1))} & \gate{R_z(\arctan(x_1^2))} & \ctrl{1}   & \qw       & \qw      & \targ   & \gate{R(\alpha_1, \beta_1, \gamma_1)} & \meter \qw \\
\lstick{\ket{0}} & \gate{R_y(\arctan(x_2))} & \gate{R_z(\arctan(x_2^2))} & \targ      & \ctrl{1}  & \qw      & \qw     & \gate{R(\alpha_2, \beta_2, \gamma_2)} &  \qw \\
\lstick{\ket{0}} & \gate{R_y(\arctan(x_3))} & \gate{R_z(\arctan(x_3^2))} & \qw        & \targ     & \ctrl{1} & \qw     & \gate{R(\alpha_3, \beta_3, \gamma_3)} &  \qw \\
\lstick{\ket{0}} & \gate{R_y(\arctan(x_4))} & \gate{R_z(\arctan(x_4^2))} & \qw        & \qw       & \targ    & \ctrl{-3}& \gate{R(\alpha_4, \beta_4, \gamma_4)} & \qw \gategroup{1}{4}{4}{8}{.7em}{--}\qw 
}
\end{minipage}
\end{center}
\caption[Variational quantum circuit component for QCNN kernel (filter).]{{\bfseries Variational quantum circuit component for QCNN kernel (filter).}
  The QCNN kernel (filter) includes three components: \emph{encoding}, \emph{variational} and \emph{quantum measurement}. The encoding component consists of several single-qubit gates $R_y(\arctan(x_i))$ and $R_z(\arctan(x_i^2))$ which represent rotations along $y$-axis and $z$-axis by the given angle $\arctan(x_i)$ and $\arctan(x_i^2)$, respectively. These rotation angles are derived from the input pixel values $x_i$ and are not subject to iterative optimization. The choose of arc tangent function is that in general the input values are not in the interval of $[-1, 1]$. 
  The variational component consists of CNOT gates between each pair of neighbouring qubits which are used to entangle quantum states from each qubit and general single qubit unitary gates $R(\alpha,\beta,\gamma)$ with three parameters $\alpha,\beta,\gamma$. Parameters labeled $\alpha_i$, $\beta_i$ and $\gamma_i$ are the ones for iterative optimization. The quantum measurement component will output the Pauli-$Z$ expectation values of designated qubits. 
  The number of qubits and the number of measurements can be adjusted to fit the problem of interest. In this work, we use the VQC as a convolutional kernel (filter), therefore the number of qubits equals to the square of kernel (filter) size and we only consider the measurement on the first qubit. The grouped box in the VQC may repeat several times to increase the number of parameters, subject to the capacity and capability of the available quantum computers or simulation software used for the experiments.}

\label{Fig:Basic_VQC}
\end{figure}
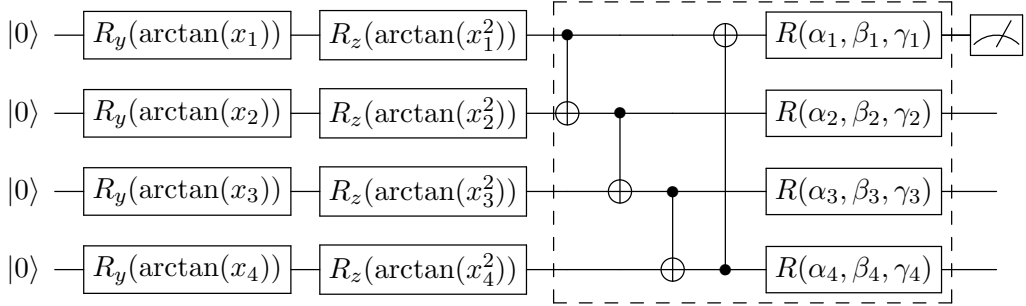

Both of the classical and quantum convolutional operations follow the following rule:
\begin{equation}
    W_{out}=\frac{\left(W_{in}-F+2 P\right)}{S}+1,
\end{equation}
where
\begin{itemize}
    \item $W_{out}$ : the output dimension of the convolutional layer
    \item $W_{in}$ : the output dimension of the convolutional layer
    \item $F$ : filter size
    \item $P$ : padding size.
\end{itemize}
To capture the spatial dependency of the input data (e.g., images), the convolutional filter will sweep across the pixels and output the corresponding values on each location (see~\figureautorefname{\ref{QCNN_Concept}}). In the QCNN, the filter itself is a VQC, which will transform an $n \times n$ dimensional vector into a single value (see ~\figureautorefname{\ref{QCNN_Operation}}).
The circuit component for the quantum convolutional filter is in the~\figureautorefname{\ref{Fig:Basic_VQC}}.
In general, each of the quantum convolutional filters captures a single kind of feature. We may place several filters in a convolutional layer to extract multiple features. In addition, we can \emph{stack} multiple convolutional layers to extract different levels of features.


\subsection{Classical Post-processing}
The output from the last quantum convolutional layer then will be flattened and processed by a single layer of a fully connected classical neural network. To represent the output values as the probabilities of each class label, we further employ the softmax function on the post-processed output.

\subsection{Loss Function and Optimization}
In this classification task, we use \emph{cross-entropy} loss, which can be written in the following formulation:
\begin{equation}
    L(\hat{\bm{y}}, \bm{y}) = -\sum_{c=1}^{M} y_{o, c} \log \left(\hat{y}_{o, c}\right),
\end{equation}
where
\begin{itemize}
    \item $M$ - the number of classes
    \item $log$ - the natural log
    \item $y_{o, c}$ - the binary indicator ($0$ or $1$) if class label $c$ is the correct classification for observation $o$
    \item $\hat{y}_{o, c}$ - predicted probability observation $o$ is of class $c$.
\end{itemize}

In this work, we use the gradient-based method to update the circuit parameters. The first problem is to calculate the gradients of quantum functions. The quantum functions are a series of operations with quantum gates, which are not the same as the layer operations in classical DNNs. In addition, the quantum functions typically are measured to retrieve the expectation values, which are stochastic by nature. In our work, we adopt the \emph{parameter-shift} rule~\cite{schuld2019evaluating, bergholm2018pennylane} to perform all of the quantum gradient calculation. 
For example, if we know how to calculate the expectation value of an observable $\hat{P}$ on our quantum function,
\begin{equation}
f\left(x ; \theta_{i}\right)=\left\langle 0\left|U_{0}^{\dagger}(x) U_{i}^{\dagger}\left(\theta_{i}\right) \hat{P} U_{i}\left(\theta_{i}\right) U_{0}(x)\right| 0\right\rangle=\left\langle x\left|U_{i}^{\dagger}\left(\theta_{i}\right) \hat{P} U_{i}\left(\theta_{i}\right)\right| x\right\rangle,
\end{equation}
where $x$ is the input value (e.g., pixel values); $U_0(x)$ is the state preparation routine to transform or encode $x$ into a quantum state; $i$ is the circuit parameter index for which the gradient is to be evaluated; and $U_i(\theta_i)$ represents the single-qubit rotation generated by the Pauli operators $X, Y$, and $Z$. It can be shown \cite{mitarai2018quantum} that the gradient of this quantum function $f$ with respect to the parameter $\theta_i$ is
\begin{equation}
    \nabla_{\theta_i} f(x;\theta_i) = \frac{1}{2}\left[ f\left(x;\theta_i + \frac{\pi}{2}\right) - f\left(x;\theta_i - \frac{\pi}{2}\right)\right].
    \label{eq:quantum gradient}
\end{equation}

%
%
Here, we have the recipe to calculate the quantum gradients. However, it still is not clear how to update the circuit parameters. In the simplest form of the gradient-descent method, the parameters are updated according to:
\begin{equation}
    \theta \leftarrow \theta - \eta \nabla_{\theta} L(x;\theta),
\end{equation}
where the $\theta$ is the model parameter, $L$ is the loss function, and $\eta$ is the learning rate. However, this vanilla form does not always work. For example, it may be easily stuck in local optimum \cite{ruder2016overview}, or it can make the model difficult to train. There are several gradient-descent variants that are successful \cite{ruder2016overview, Tieleman2012, kingma2014adam}. Based on previous works \cite{chen2020quantum, chen19}, we use the RMSProp optimizer to optimize our hybrid quantum-classical model.
RMSProp \cite{Tieleman2012} is a special kind of gradient-descent method with an adaptive learning rate that updates the parameters $\theta$ as:
\begin{subequations}
\begin{align}
        E\left[g^{2}\right]_{t} &= \alpha E\left[g^{2}\right]_{t-1}+ (1 - \alpha) g_{t}^{2}, \\ 
        \theta_{t+1} &= \theta_{t}-\frac{\eta}{\sqrt{E\left[g^{2}\right]_{t}}+\epsilon} g_{t},
\end{align}
\end{subequations} 
where $g_t$ is the gradient at step $t$ and $E\left[g^{2}\right]_{t}$ is the weighted moving average of the squared gradient with $E[g^2]_{t=0} = g_0^2$. The hyperparameters are set for all experiments in this paper as follows: learning rate $\eta =0.01$, smoothing constant $\alpha = 0.99$, and $\epsilon = 10^{-8}$.

%
\subsection{Dropout}
\emph{Overfitting} is a phenomenon where a machine learning model learned the statistical noise in the training data. This will cause poor performance when the models are tested against the unseen data or testing data. In other words, the model is not well generalizable. Such difficulties often emerge when training a classical DNN on a relatively small training set, and a QCNN is no exception.

One potential method to reduce overfitting is to train all possible neural network architectures on the given dataset and average the predictions from each model. However, this is impractical because it will require unlimited computational resources. \emph{Dropout} is a method that approximates the effect of training a large number of neural networks with different architectures simultaneously~\cite{srivastava2014dropout}. 

The dropout operation entails (as follows): during the \emph{training} phase, on each of the forward passes, some of the output values from the specified layer will become zeroes. Each one of the values from the specified layer will be zeroed independently with probability $p$ from a Bernoulli distribution. This dropout procedure will not be performed in the \emph{testing} phase.
\section{Experiments and Results}\label{sec:results}
\begin{figure}[htbp]
\centering
\includegraphics[width=1.\linewidth]{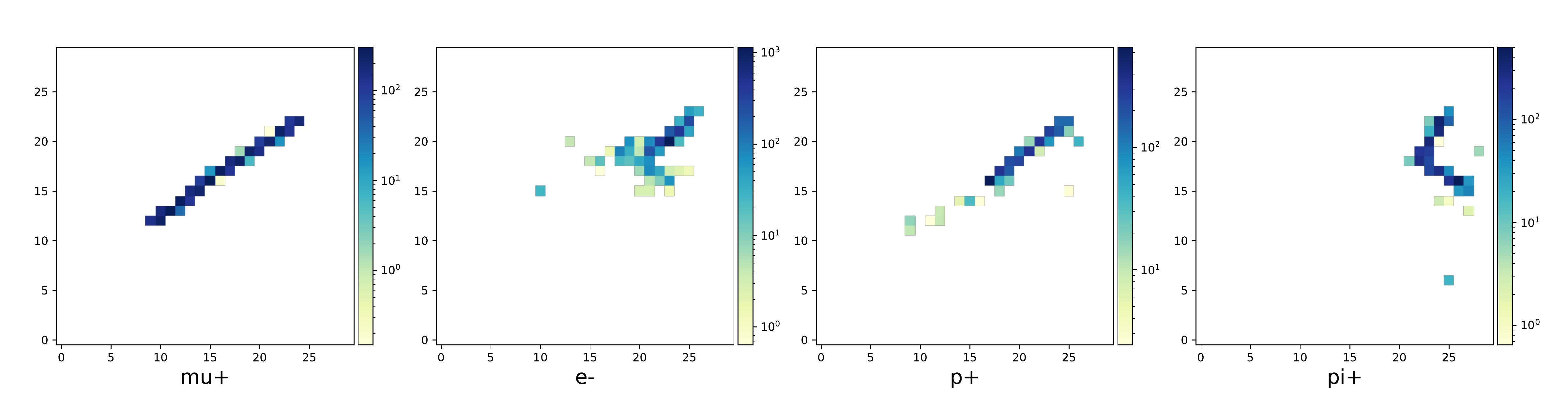}
\caption{Examples of scaled images of simulated particle activities ($\mu$, $\pi^+$, $p$, $e^-$) in a LArTPC detector. These are the images used in the QCNN experiments. The dimension of these images is $30 \times 30$ pixels.}
\label{scaled_examples}
\end{figure}

This section presents the numerical simulation of QCNN on the task of classifying different HEP events. The input data are in the dimension of $30 \times 30$. To demonstrate the possible quantum advantage, the classical and quantum CNN have a comparable number of parameters. \figureautorefname{\ref{scaled_examples}} shows the examples of the data used to train and test the QCNN models.
For fair comparison, we arrange the experiment of QCNN and CNN to have similar numbers of parameters. In the classical CNN part, there are $4$ channels in the first convolutional layer with the filter size as $5 \times 5$ and $2$ channels in the second convolutional layer with the filter size as $5 \times 5$. Finally, there is a fully connected layer, which features $7 \times  7 \times 2 \times 2 + 2 = 198$ parameters. Therefore, the total number of parameters in the classical CNN is $4 \times 5 \times 5 + 4 \times 2 \times 5 \times 5 + 198 = 498$. For the QCNN, there is $1$ channel in the first quantum convolutional layer with the filter size of $3 \times 3$ and another single channel in the second quantum convolutional layer with the filter size of $2 \times 2$. Finally, there is a classical fully connected layer with $ 14 \times 14 \times 1 \times 2 + 2 = 394$ parameters. As such, the total number of the hybrid quantum-classical CNN is $54 + 24 + 394 = 472$. 
The software used for this work are PyTorch \cite{paszke2019pytorch}, PennyLane \cite{bergholm2018pennylane}, and Qulacs \cite{suzuki2020qulacs}.
%

%

%
%
\subsection{Muon versus Electron}
\begin{figure}[htbp]
\centering
\includegraphics[width=1.0\linewidth]{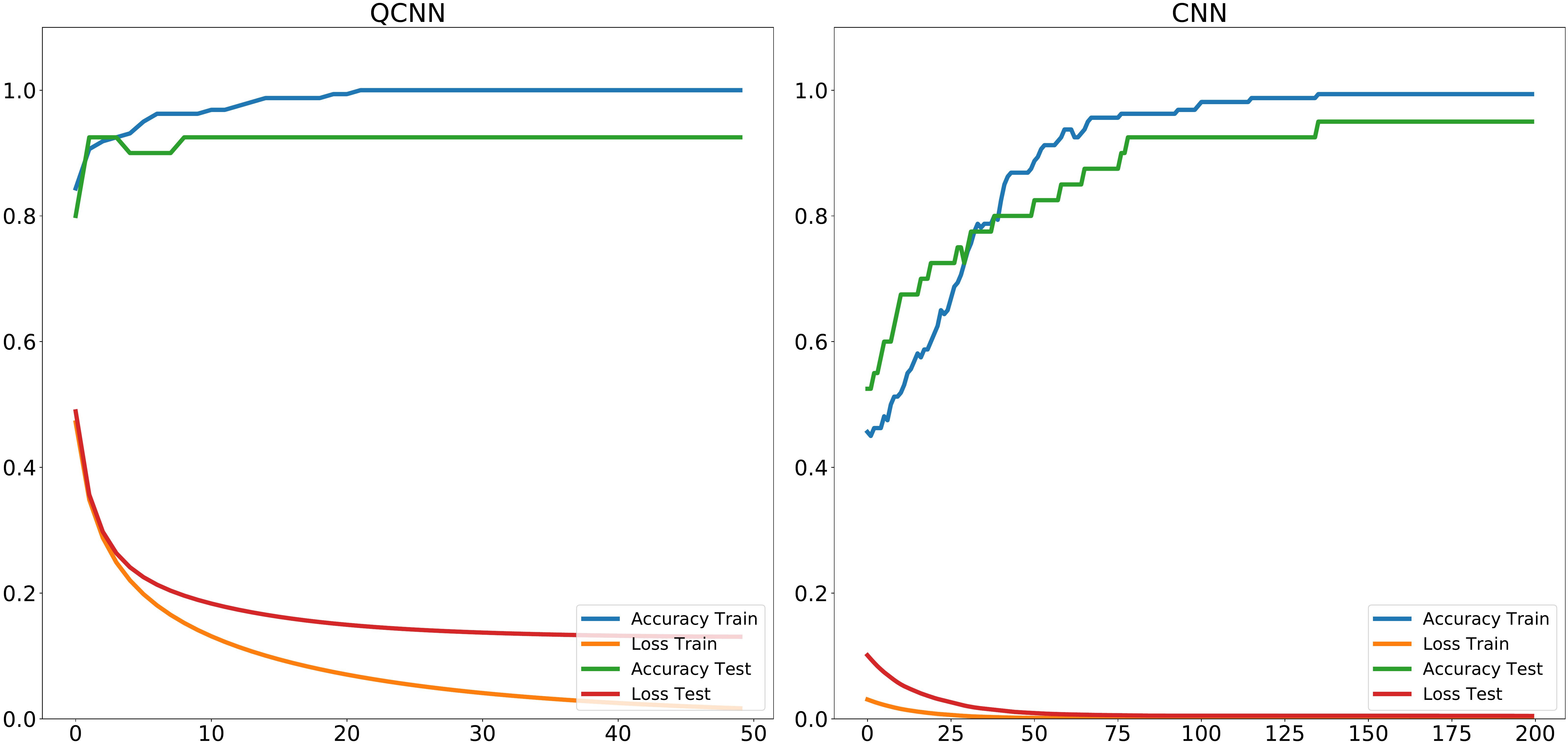}
\caption[Result: QCNN on binary classification of muon versus electron]{{\bfseries Result: QCNN on binary classification of muon versus electron}
Training QCNN on the classification of $\mu^+$ and $e^-$. The filter size is $3$ in the first convolutional layer and $2$ in the second convolutional layer. There is $1$ channel in both Conv Layers. The number of parameters in this setting: $9 \times 3 \times 2 = 54$ in the first convolutional layer, $4 \times 3 \times 2 = 24$ in the second convolutional layer, and $14 \times 14 \times 1 \times 2 + 2= 394$ in the fully connected layer. Total number of parameters is $54 + 24 + 394 = 472$.}
\label{comparison_results_mu_electron}
\end{figure}

\figureautorefname{\ref{comparison_results_mu_electron}} and Table~\ref{tab:results_comparison_mu_e} depict the results of the classification between $\mu^+$ and $e^-$. As mentioned in Section~\ref{sec:hep-data}, A $\mu^+$ is a track-like particle, while an $e^-$ produces electromagnetic showers that are spatially extended. This is a relatively straightforward pattern recognition problem when the particle tracks are more than a few meters ($\sim$2 meters in this simulation). In fact, we see that the test accuracy in QCNN (92.5\%) and CNN (95\%) is comparable to each other with a comparable number of parameters. On the other hand, the QCNN converges to its optimal accuracy much faster with a fewer number of epochs.

\begin{table}[htbp]
\centering
\begin{tabular}{|l|l|l|l|l|}
\hline
     & Training Accuracy & Testing Accuracy & Training Loss & Testing Loss \\ \hline
QCNN & $100\%$              & $92.5\%$             & $0.017$            & $0.13$           \\ \hline
CNN  & $99.38\%$              & $95\%$             & $0.0002$            & $0.0046$           \\ \hline
\end{tabular}
\caption{Performance comparison between QCNN and CNN on the binary classification between $\mu^+$ and $e^-$.}
\label{tab:results_comparison_mu_e}
\end{table}

\subsection{Muon versus Proton}
\begin{figure}[htbp]
\centering
\includegraphics[width=1.\linewidth]{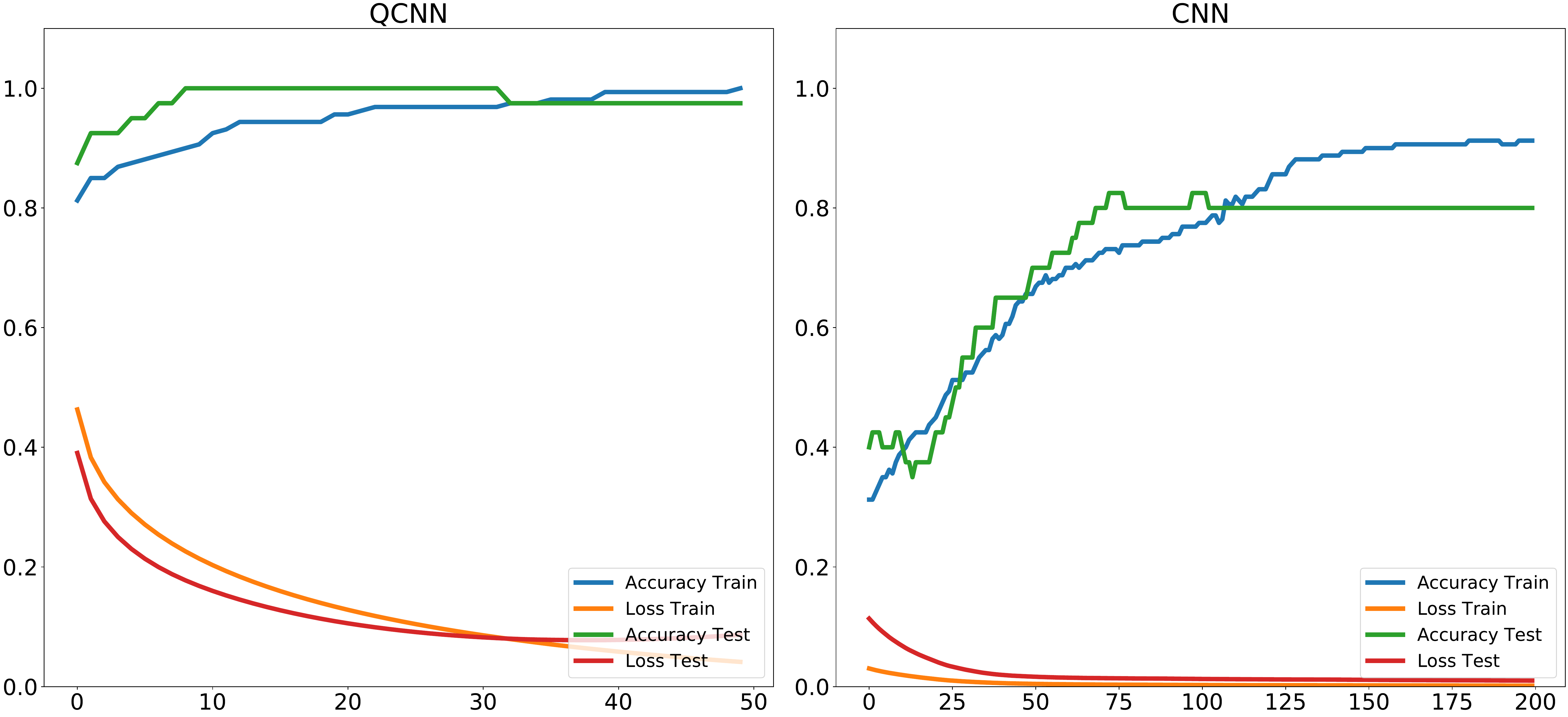}
\caption[Result: QCNN on binary classification of muon versus proton]{{\bfseries Result: QCNN on binary classification of muon versus proton}
Training QCNN on the classification of $\mu^+$ and $p$. The filter size is $3$ in the first convolutional layer and $2$ in the second convolutional layer. There is $1$ channel in both Conv Layers. The number of parameters in this setting: $9 \times 3 \times 2 = 54$ in the first convolutional layer, $4 \times 3 \times 2 = 24$ in the second convolutional layer, and $14 \times 14 \times 1 \times 2 + 2= 394$ in the fully connected layer. Total number of parameters is $54 + 24 + 394 = 472$.}
\label{comparison_results_mu_proton}
\end{figure}

\figureautorefname{\ref{comparison_results_mu_proton}} and Table~\ref{tab:results_comparison_mu_p} show the results of the classification between $\mu^+$ and $p$. As described in Section~\ref{sec:hep-data}, a proton is a track-like particle akin to a muon. However, because a proton's mass is much heavier than a muon or pion, it has higher energy loss and encounters less MCS when it passes the detector. As a result, a proton's track has higher intensity and is straighter than that of a muon. This classification is more difficult than the previous case detailing muon versus electron, which is evident from the CNN's test accuracy of 80\%. In this case, we show that with a comparable number of parameters, the QCNN outperforms the classical CNN, both in test accuracy and learning speed. The QCNN reaches $97.5$\% test accuracy at around $10$ epochs, while the classical CNN plateaus at a test accuracy of $80\%$ at roughly $75$ epochs. 
\begin{table}[htbp]
\centering
\begin{tabular}{|l|l|l|l|l|}
\hline
     & Training Accuracy & Testing Accuracy & Training Loss & Testing Loss \\ \hline
QCNN & $100.00\%$        & $97.5\%$             & $0.041$       & $0.087$           \\ \hline
CNN  & $91.25\%$         & $80\%$             & $0.002$            & $0.01$           \\ \hline
\end{tabular}
\caption{Performance comparison between QCNN and CNN on the binary classification between $\mu^+$ and $p$.}
\label{tab:results_comparison_mu_p}
\end{table}

\subsection{Muon versus Charged Pion}
\begin{figure}[htbp]
\centering
\includegraphics[width=1.\linewidth]{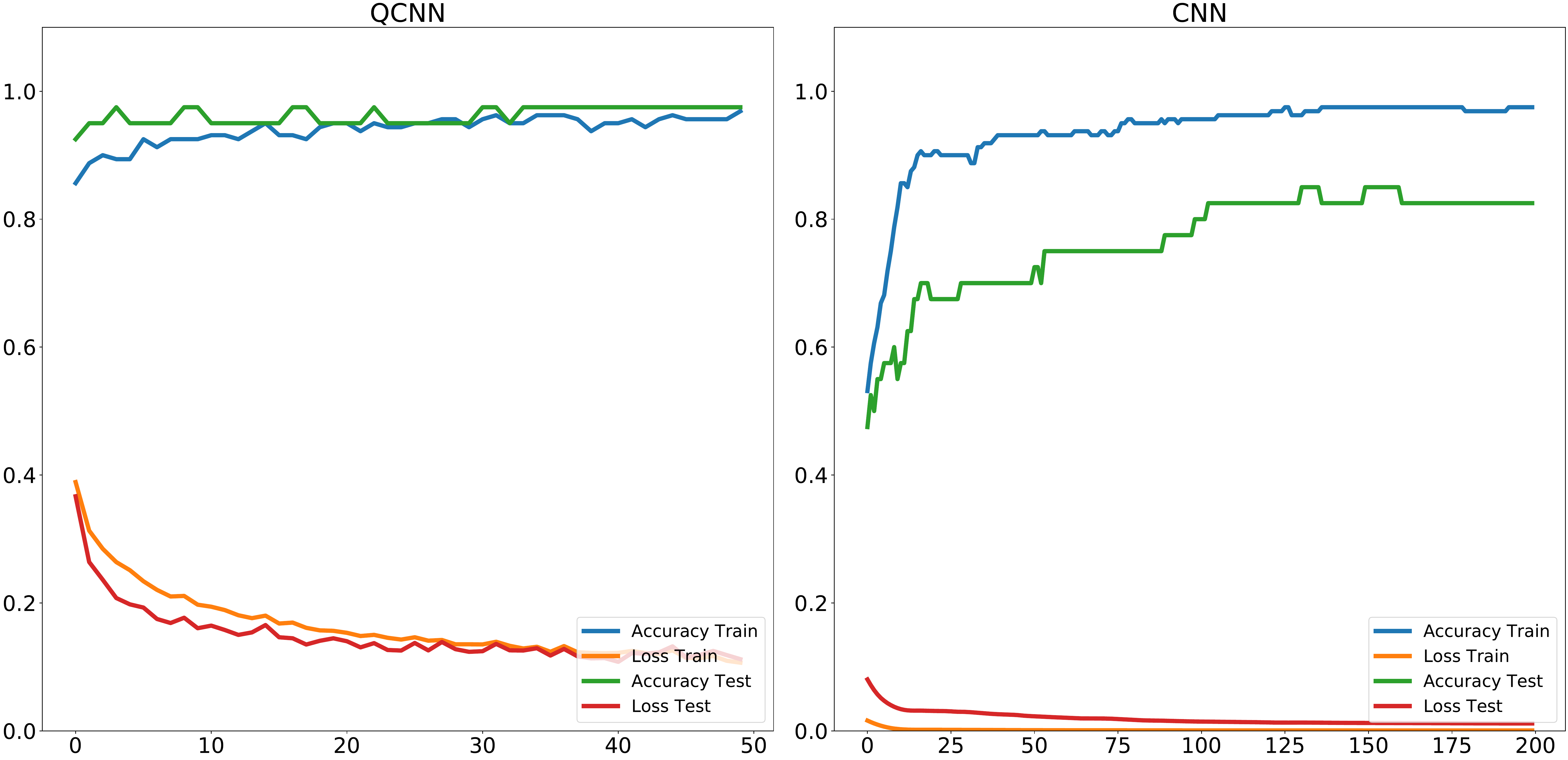}
\caption[Result: QCNN on binary classification of muon versus charged pion]{{\bfseries Result: QCNN on binary classification of muon versus charged pion}
Training QCNN on the classification of $\mu^+$ and $\pi^+$. The filter size is $3$ in the first convolutional layer and $2$ in the second convolutional layer. There is $1$ channel in both Conv Layers. The number of parameters in this setting: $9 \times 3 \times 2 = 54$ in the first convolutional layer, $4 \times 3 \times 2 = 24$ in the second convolutional layer, and $14 \times 14 \times 1 \times 2 + 2= 394$ in the fully connected layer. Total number of parameters is $54 + 24 + 394 = 472$. In this experiment, we add a dropout layer with dropout rate of $0.3$ to the QCNN.}
\label{comparison_results_mu_charged_pion}
\end{figure}

\figureautorefname{\ref{comparison_results_mu_charged_pion}} and Table~\ref{tab:results_comparison_mu_pi} illustrate the results of the classification between $\mu^+$ and $\pi^+$. This is another difficult classification problem because a charged pion behaves much like a muon in terms of energy loss, MCS, and decay.  As described in Section~\ref{sec:hep-data}, the main difference is that the $\pi^+$ experiences additional nuclear interactions during its passage in the detector, often leading to a hard scattering (represented as a ``kink'') along its main trajectory. In this case, we show that with a comparable number of parameters, the QCNN outperforms the classical CNN, both in test accuracy and learning speed. The QCNN reaches $97.5$\% test accuracy at the first few epochs, while the classical CNN flattens at the test accuracy of $82.5\%$ at around $100$ epochs. In this case, we add a dropout layer in the QCNN with dropout rate $=0.3$ to improve its robustness against overfitting.

\begin{table}[htbp]
\centering
\begin{tabular}{|l|l|l|l|l|}
\hline
     & Training Accuracy & Testing Accuracy & Training Loss & Testing Loss \\ \hline
QCNN & $96.88\%$         & $97.5\%$         & $0.1066$      & $0.1121$     \\ \hline
CNN  & $97.5\%$              & $82.5\%$             & $0.0006$            & $0.0116$           \\ \hline
\end{tabular}
\caption{Performance comparison between QCNN and CNN on the binary classification between $\mu^+$ an $\pi^+$.}
\label{tab:results_comparison_mu_pi}
\end{table}

%
%
\section{Related Works}
The concept of QCNNs has been discussed recently. In~\cite{cong2019quantum}, the authors propose an architecture based on the Multiscale Entanglement Renormalization Ansatz (MERA) tensor network to perform the classification of quantum states. Our approach differs from this work as we focus on the classical input data. In~\cite{kerenidis2019quantum, li2020quantum}, the authors propose a QCNN framework to deal with the classical data, which is like our method in the sense of targets. However, those works require the operation of quantum random access memory (QRAM), which is difficult to implement on physical devices in the near term. In~\cite{oh2020tutorial, liu2019hybrid}, the authors consider a more realistic architecture that also is hybrid quantum-classical. While in a similar vein, our work differs from that research because it implements input data with much larger dimensions. In~\cite{liu2019hybrid}, the data are in the dimension of $3 \times 3$, while in~\cite{oh2020tutorial}, the data are in the dimension of $10 \times 10$. Our architecture is capable of dealing with dimensions up to $30 \times 30$. We also noted the work in \cite{henderson2020quanvolutional}. In~\cite{henderson2020quanvolutional,yang2020decentralizing}, quantum circuits are randomly sampled and not subject to iterative optimization. In our work, the quantum and classical parts are trained in an end-to-end fashion.
The trainability of QCNN is studied in the recent work, \cite{pesah2020absence}, indicating that QCNN optimization is more viable than other quantum neural network architectures.

\section{Conclusion and Outlook} \label{sec:conclusion}

In this work, we propose a quantum machine learning framework for learning HEP events. Specifically, we demonstrate the QCNN architecture with significant learning capacity in terms of learning speed and testing accuracy compared to the classical CNN when both use a comparable number of parameters. We expect the proposed framework will have a wide range of applications in the era of Noisy Intermediate-Scale Quantum (NISQ) and beyond, as well as in more HEP experiments. 

Looking ahead, there are several research areas where we could extend our QCNN framework. First, in this work, we perform experiments using the input dimension of $30 \times 30$ and a single input channel. In comparison, multiple input channels (e.g., different color channels) with higher dimensions are rather common in a classical CNN. Future studies to increase the data complexity in QCNNs are expected as the speed of quantum simulators improves. Second, this work presents a noise-free simulation to demonstrate proof-of-concept QCNN experiments on HEP event classification. Our framework is based on VQC, which has the potential to be robust against device noise. However, applying parameter-shift methods to calculate quantum gradients requires a large amount of quantum circuit evaluations, which is infeasible at this time. For example, given a filter with the size $N \times N$ and the number of operations needed to scan over a single input images $S$, the contribution to the total number of circuit evaluation is, at least, $\mathcal{O}(N^2S)$. Therefore, the number of total circuit evaluations grows as the circuit goes deeper (with more convolutional layers) or expands wider (with more filters in each layer). We reserve pursuing this area of study until the appropriate quantum computing resources are available. Finally, as the convolutional operation is quite versatile, it has been widely used beyond computer vision in classical machine learning, for example, in modeling data with temporal or sequential dependencies \cite{kalchbrenner2014convolutional, hu2014convolutional, acharya2017deep, lai2015recurrent, yin2016abcnn}. As such, our proposed general QCNN architecture is not limited to image classification tasks and can be extended to other application domains. 

%

\begin{acknowledgments}
This work is supported by the U.S.\ Department of Energy, Office of Science, Office of High Energy Physics program under Award Number DE-SC-0012704 and the Brookhaven National Laboratory Directed Research and Development (LDRD) Program \#20-024.
\end{acknowledgments}

\appendix


\bibliographystyle{ieeetr}
\bibliography{apssamp,bib/tool,bib/qcnn_related,bib/qml_examples,bib/classical_cnn,bib/classical_cnn_for_science,bib/nisq,bib/classical_cnn_beyond_cv,bib/qc}

\end{document}